\pdfoutput=1
\documentclass[11pt]{article}

\usepackage{ACL2023}

\usepackage{times}
\usepackage{latexsym}

\usepackage[T1]{fontenc}
\usepackage[utf8]{inputenc}

\usepackage{microtype}

\usepackage{inconsolata}

\usepackage{graphicx}
\usepackage{subfigure}
\usepackage{amsmath}
\usepackage{amssymb}
\usepackage{booktabs}
\usepackage{stfloats}
\usepackage{float}
\graphicspath{{figure/}}
\usepackage{algorithm}
\usepackage{algorithmic}
\usepackage{bbding}

\title{HiFi: High-Information Attention Heads Hold for Parameter-Efficient Model Adaptation}
\author{
    Anchun Gui and Han Xiao\thanks{~~Corresponding author.} \\
    Department of Artificial Intelligence \\
    School of Informatics, Xiamen University \\
    \texttt{anchungui@stu.xmu.edu.cn, bookman@xmu.edu.cn}
}
\begin{document}
\maketitle

\begin{abstract}
To fully leverage the advantages of large-scale pre-trained language models (PLMs) on downstream tasks, it has become a ubiquitous adaptation paradigm to fine-tune the entire parameters of PLMs. However, this paradigm poses issues of inefficient updating and resource over-consuming for fine-tuning in data-scarce and resource-limited scenarios, because of the large scale of parameters in PLMs. To alleviate these concerns, in this paper, we propose a parameter-efficient fine-tuning method HiFi, that is, only the highly informative and strongly correlated attention heads for the specific task are fine-tuned. To search for those significant attention heads, we develop a novel framework to analyze the effectiveness of heads. Specifically, we first model the relationship between heads into a graph from two perspectives of information richness and correlation, and then apply PageRank algorithm to determine the relative importance of each head. Extensive experiments on the GLUE benchmark demonstrate the effectiveness of our method, and show that HiFi obtains state-of-the-art performance over the prior baselines.
\end{abstract}

\section{Introduction}
\label{sec:intro}
Recently large-scale pre-trained language models (PLMs) have triggered a technological revolution in natural language processing (NLP), as the satisfactory performance could be achieved by fully fine-tuning parameters of PLMs \cite{devlin2019bert, brown2020language, wei2021nezha}.
In data-scarce and resource-limited scenarios, however, this methodology poses several concerns.
On one hand, full fine-tuning leads to inefficient updating and catastrophic forgetting issues when the training set is insufficient \cite{houlsby2019parameterefficient, wang2021kadapter}; on the other hand, this approach has to duplicate a modified copy of full parameters per task, which presents the challenge of resource over-consuming for the limited storage infrastructure.

\begin{figure}[!t]
    \centering
    \includegraphics[width=0.4\textwidth]{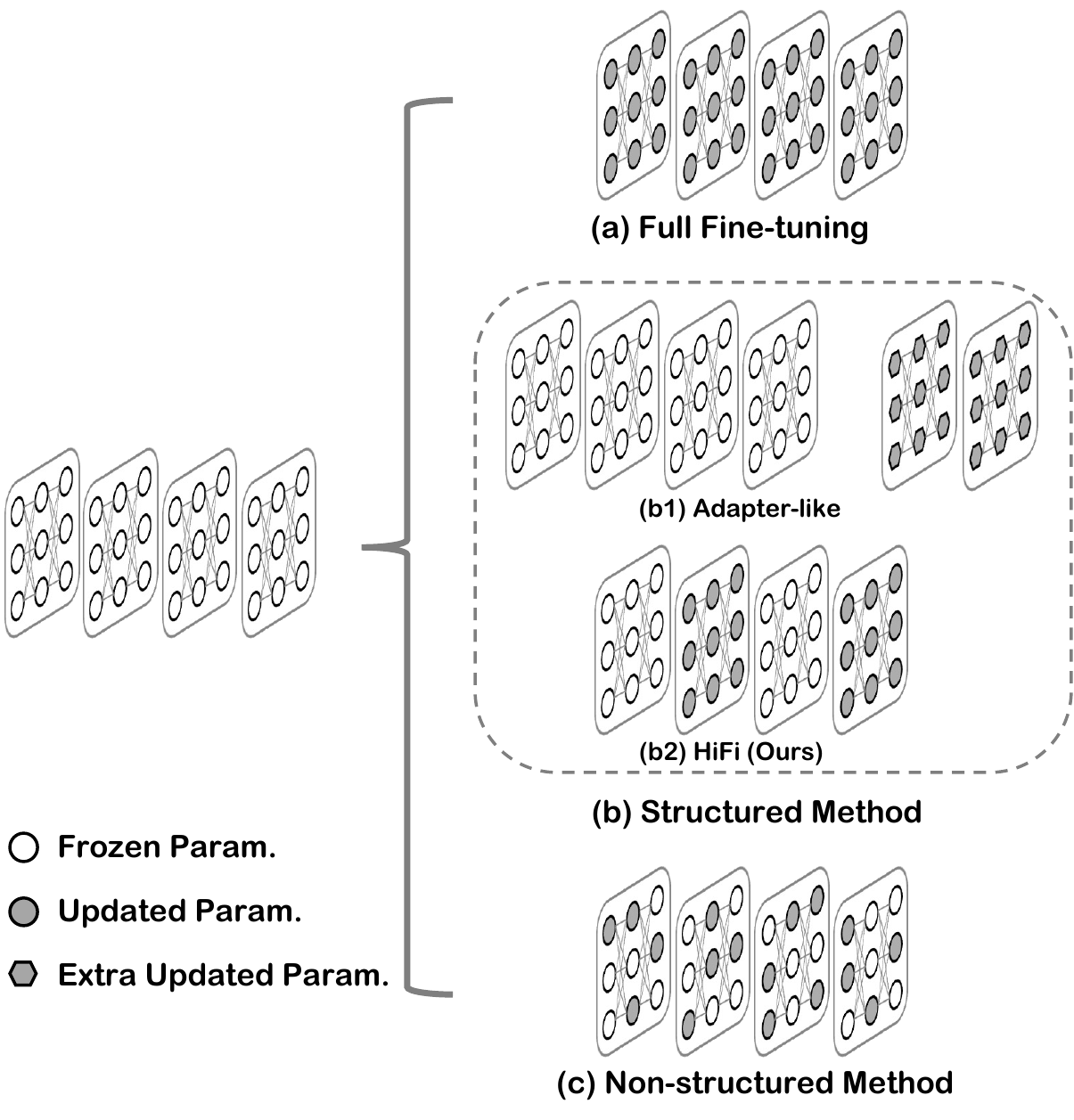}
    \caption{Comparison of diverse fine-tuning paradigms. For an example of a set of head weights, full fine-tuning updates all weights, whereas the non-structured methods randomly fine-tune a subset of parameters. For the structured methods, Adapter-like models update the extra weights, while our proposed HiFi selects several vital heads for fine-tuning.}
    \label{fig:intro}
\end{figure}

Parameter-efficient fine-tuning (PEFT), as an alternative paradigm, has attracted more attention recently \cite{he2022unified, ding2022delta}. Compared to full fine-tuning, PEFT only fine-tunes the minority of the original parameters (or extra introduced parameters) instead of the entire parameters of PLMs, as shown in Fig.~\ref{fig:intro}. Although current PEFTs effectively decrease the proportion of trainable parameters, these methods also lead to varying aspects of concerns. For instance, Adapter \cite{houlsby2019parameterefficient} not only breaks the model structure by introducing additional parameters but also causes inference delays \cite{hu2022lora}. We compare the representative models in Tab.~\ref{tab:comparison}.

Motivated by these issues, in this paper, we propose HiFi, a novel PEFT method by fine-tuning the relatively significant heads in multi-head attention (MHA) modules, where we assume that our PLMs are based on Transformer \cite{vaswani2017attention} architecture and MHA is chosen since it plays a crucial role in Transformer according to recent studies \cite{voita2019analyzing, baan2019understanding, michel2019are}. There are several intractable challenges to implementing our proposed method.

\textbf{How to measure the individual importance of a head?}
The core question of HiFi is to select relative importance heads in MHA. Toward this end, we first analyze the importance of a single head. Specifically, we decompose the output of each head by singular value decomposition (SVD) to obtain a sequence of singular values in descending order, where if the cumulation of top-$t$ terms account for a percentage threshold (e.g., $90\%$), the index $t$ is treated as a measurement of information richness of the head.

\begin{figure*}[t]
    \centering
    \includegraphics[width=1.0\textwidth]{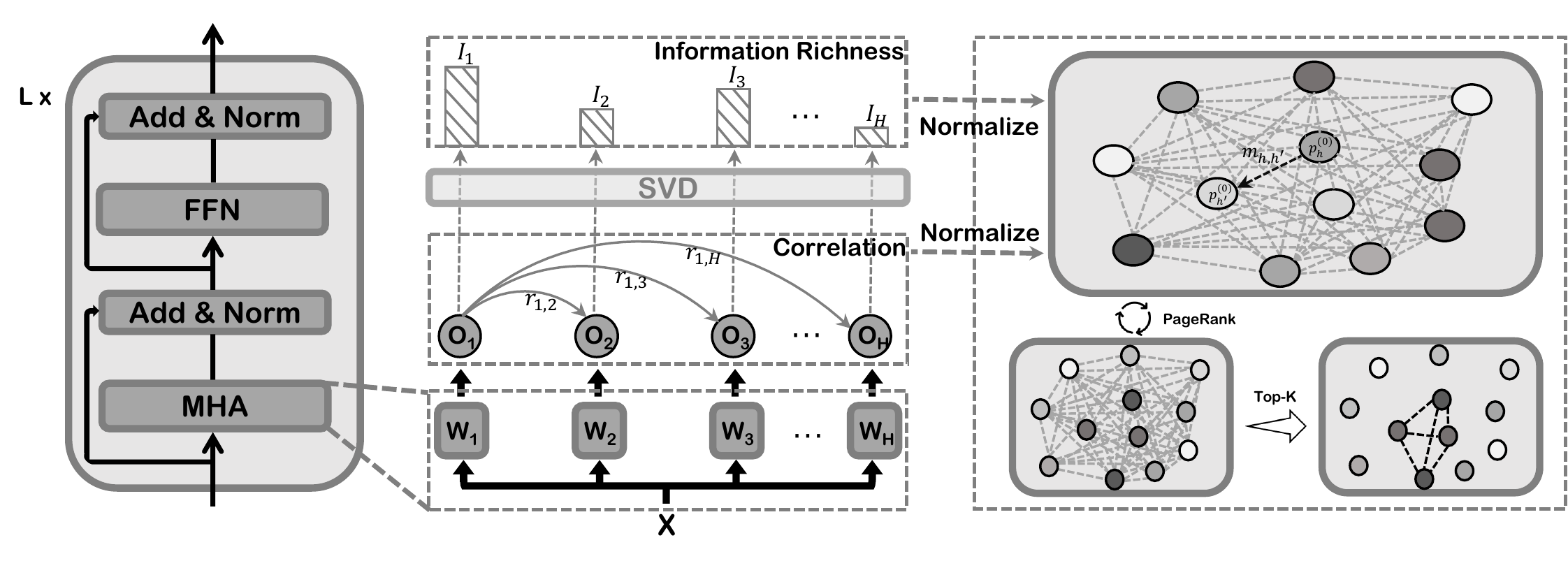}
    \caption{An overview of our method. For each layer, we first calculate the information richness of a single head and the correlation between heads, then construct a graph by normalizing our proposed metrics. For a specific downstream task, we search for the relative significant heads for fine-tuning using PageRank algorithm. The darker ball on the right figure indicates more important head.}
    \label{fig:overview}
\end{figure*}

\textbf{How to measure the relative importance between heads?}
Given that the collaboration among multiple heads achieves success \cite{kovaleva2019revealing, clark2019what}, the head-to-head correlation also shall be considered. Therefore, we calculate the covariance matrix between the outputs of heads to measure the correlation across heads. To further work out those highly informative and strongly correlated heads, we model the relationship between heads into a graph based on their information richness and correlation, and then derive the relative importance of each head by PageRank algorithm \cite{page1999pagerank}. We illustrate our method overview in Fig.~\ref{fig:overview}. 

To verify the effectiveness of our proposed method, we conduct extensive experiments on the GLUE benchmark \cite{wang2018glue}, in both full-shot\footnote{It refers to vanilla fine-tuning setting, see Sec.~\ref{sub_sec:setup}.} and few-shot settings. The results show that our model HiFi gains state-of-the-art performance against strong baselines. For example, full fine-tuning achieves the average score of $82.0\%$ in the full-shot setting, while HiFi obtains superior performance ($82.3\%$).

To summarize, our contributions are as follows:
\begin{itemize}
    \item We develop a novel framework for analyzing the relative importance of weights, and empirically demonstrate its robustness under diverse experimental settings.
    \item Based on this framework, we propose a simple yet effective PEFT method, HiFi. Our method fulfills the performance requirement without introducing additional concerns compared to the previous baselines, as shown in Tab.~\ref{tab:comparison}.
    \item Our proposed HiFi outperforms the current strong counterparts on the GLUE benchmark, in both full-shot and few-shot settings. We also verify the effectiveness of our methodology by abundant analytical experiments.
\end{itemize}

\begin{table}[!t]
    \centering
    \scalebox{0.63}{
        \begin{tabular}{l | c c c c}
            \toprule
                Method & Extra Param. & Corrupt Struc. & Infer. Delay & Store Cons. \\
            \midrule
                Full-FT & \color{red}\XSolidBrush & \color{red}\XSolidBrush & \color{red}\XSolidBrush & \Checkmark \\
                Adapter & \Checkmark & \Checkmark & \Checkmark & \color{red}\XSolidBrush \\
                LoRA & \Checkmark & \Checkmark & \color{red}\XSolidBrush & \color{red}\XSolidBrush \\
                Prompt-Tuning & \Checkmark & \color{red}\XSolidBrush & \Checkmark & \color{red}\XSolidBrush \\
                Diff-Pruning & \Checkmark & \color{red}\XSolidBrush & \color{red}\XSolidBrush & \Checkmark \\
                Child-Tuning & \color{red}\XSolidBrush & \color{red}\XSolidBrush & \color{red}\XSolidBrush & \Checkmark \\
            \midrule
                \textbf{HiFi (Ours)} & \color{red}\XSolidBrush & \color{red}\XSolidBrush & \color{red}\XSolidBrush & \color{red}\XSolidBrush \\
            \bottomrule
        \end{tabular}
    }
    \caption{Compared with related methods, our proposed HiFi does not raise additional concerns. ``Extra Param.'': introduces new trainable parameters apart from the original parameters. ``Corrupt Struc.'': corrupts the original structure of the model. ``Infer. Delay'': causes inference delay. ``Store Cons.'': saves all parameters per task or the updated parameters are not convenient to store\protect\footnotemark.}
    \label{tab:comparison}
\end{table}
\footnotetext{For the non-structured methods (e.g., Diff-Pruning \cite{guo2021parameterefficient}, Child-Tuning \cite{xu2021raise}), since the position of updated parameters is unordered as shown in Fig.~\ref{fig:intro}(c), we cannot directly save them in a weight matrix.}

\section{Related Work}
\label{sec:related_work}
The existent research on PEFT can be generally divided into two folds: structured and non-structured methods. For the structured methods, the updated parameters are modularized (i.e., the parameters from the particular weight blocks are tuned), while the location of those updated parameters in the non-structured methods is irregular, as shown in Fig.~\ref{fig:intro}.

\paragraph{\textbf{Structured Methods.}} There are two types of updatable blocks.
(\romannumeral 1) Extra introduced blocks. For example, Adapter \cite{houlsby2019parameterefficient} inserts compact modules into PLMs, and LoRA \cite{hu2022lora} introduces two low-rank MLPs into the query and key weights in MHA. Similar models include Compacter/Compacter++ \cite{mahabadi2021compacter}, AdapterDrop \cite{ruckle2021adapterdrop}, AdapterBias \cite{fu2022adapterbias}, and MAM \cite{he2022unified}. In addition, Prompt-Tuning \cite{li2021prefixtuning, lester2021power, liu2022ptuning} is also a popular research direction via prepending a sequence of continuous learnable vectors to the input. Generally, we refer these models to Adapter-like methods in Fig.~\ref{fig:intro}, because they introduce additional learnable parameters apart from the original model parameters.
(\romannumeral 2) Internal original blocks. BitFit \cite{benzaken2022bitfit}, for instance, fine-tunes the all bias terms of PLMs on downstream tasks. The distinctions between our proposed HiFi and BitFit are as follows: 
1) we tune the attention heads rather than bias, given that the heads act a significant role in Transformer \cite{voita2019analyzing, baan2019understanding};
2) our selected heads are task-specific, while BitFit does not consider the information of downstream tasks.

\paragraph{\textbf{Non-structured Methods.}} The core question in this direction is that: how to identify a sparse sub-network based on the importance of each parameter in PLMs?
Child-Tuning \cite{xu2021raise} calculates the Fisher Information Matrix (FIM) of parameters to determine a ``child'' network in the original network. Diff-Pruning \cite{guo2021parameterefficient} learns a perturbation variable per parameter by $L_{0}$ norm constraint so that the updatable parameters are selected automatically. Similar work includes \citet{ansell2022composable}.

\section{Methodology}
\subsection{Notations}
\label{sub_sec:notations}
Supposing the PLM consists of $L$ encoder layers, and MHA has $H$ attention heads and the corresponding weights\footnote{We ignore the bias terms for simplicity.} are $W_{h}^{Q}, W_{h}^{K}, W_{h}^{V} \in \mathbb{R}^{D \times D'}, h \in \{1, 2, \cdots, H\}$, where $D$ refers to the hidden representation dimensions and $D' = \frac{D}{H}$. Let the weight set of the $h$-th head be $W_{h} = \{W_{h}^{Q}, W_{h}^{K}, W_{h}^{V}\}$. For a sample $x$ from the data distribution $\mathcal{D}$, $O_{h}(x) \in \mathbb{R}^{S \times D'}$ represents the output of $x$ through the $h$-th head, where $S$ indicates the sequence length of $x$.

Besides, in our work, we notate that $g(\cdot)$ measures the individual importance of a head and $r(\cdot, \cdot)$ indicates the correlation between two heads. The design of metrics should satisfy the following principles: task-relevant and robustness, because we expect $g(\cdot), r(\cdot, \cdot)$ to capture the intrinsic properties of heads on a range of downstream tasks.

\subsection{Information Richness}
Inspired by leveraging the feature map to measure the importance of the corresponding convolution kernel in computer vision \cite{lin2020hrank}, we here treat $O_{h}$ equivalently as the ``feature map'' in our scenario. Specifically, rather than focusing solely on $W_{h}$, we analyze its output $O_{h}$ to reflect the importance of $W_{h}$ and give the following definition:
\begin{align}
    g(W_{h}) = \mathbb{E}_{x\sim \mathcal{D}} \left[ g(O_{h}(x)) \right]
    \label{eq:g_W_h}
\end{align}
By calculating the expectation of $O_{h}$ with respect to $x$, we can measure the importance of the $h$-th head based on a particular task. The more critical the head, intuitively, the richer the corresponding output should be. To this end, we characterize this property from the perspective of singular value decomposition (SVD). Specifically, we perform the SVD on the output, i.e., $O_{h}(x) = U \Sigma V^{\top} = U~\mathrm{diag}(\sigma_{1}, \sigma_{2}, \cdots, \sigma_{T})~V^{\top}$, where $T = \min\{S, D'\}, \sigma_{1} \ge \sigma_{2} \ge \cdots \ge \sigma_{T}$.

\begin{figure*}[t]
    \centering
    \includegraphics[width=1.0\textwidth]{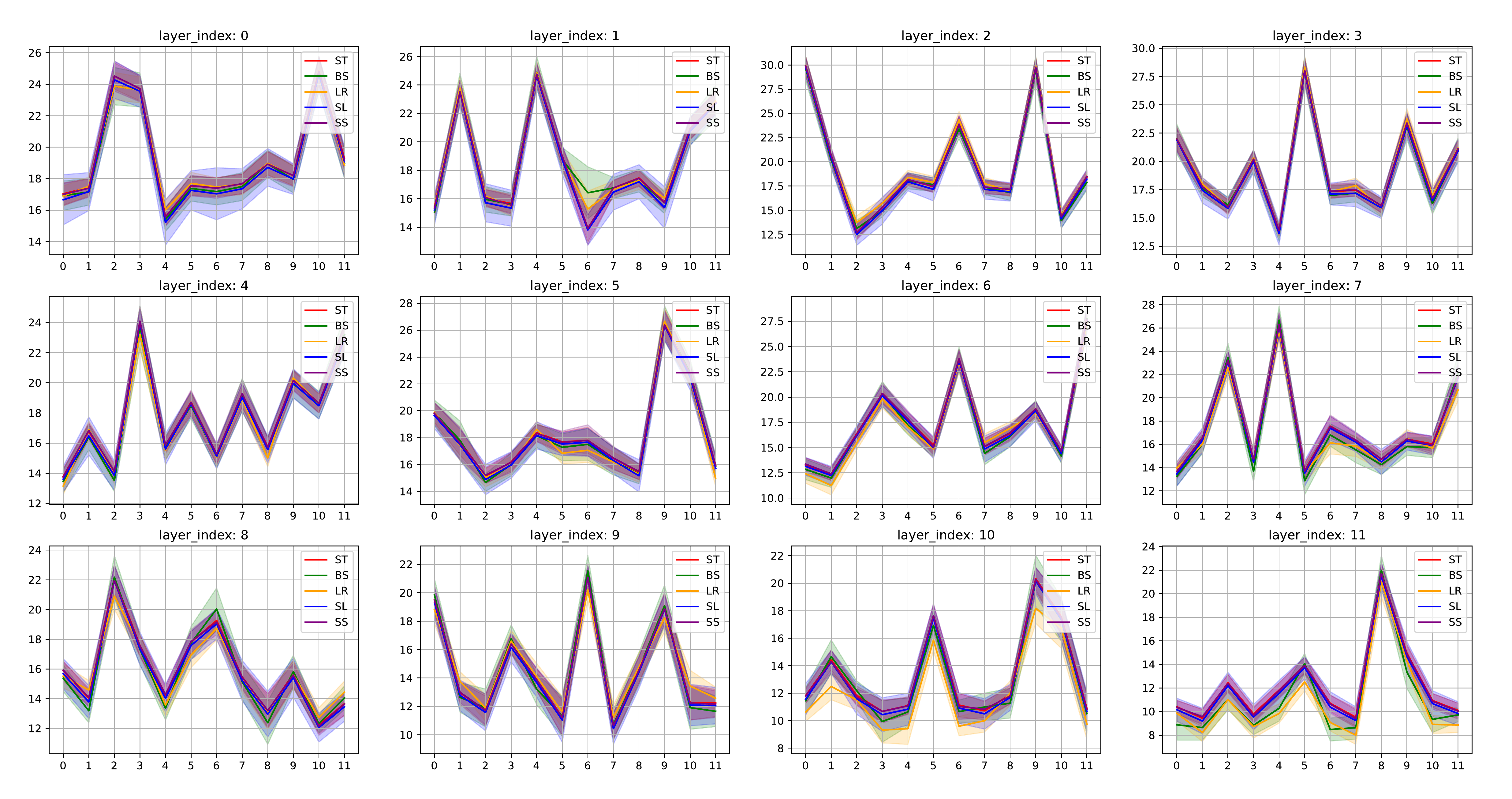}
    \caption{The robustness of information richness ($I_{h}$). Compared to the standard setting of \textbf{ST}, \textbf{BS} reduces the batch size, \textbf{LR} increases the learning rate, \textbf{SL} reduces the sequence length, and \textbf{SS} increases the sample size. In each subfigure, $x$-axis and $y$-axis represent the index of heads and corresponding $I_{h}$. The solid line and shading area refer to the mean and standard deviation, respectively. See Sec.~\ref{sub_sec:robustness} for detailed experimental settings.}
    \label{fig:info_richness}
\end{figure*}

Based on the properties of SVD, we are aware that if the sequence of singular values $\{\sigma_{t}\}$ decays slower, it means that $O_{h}$ is informative and contains more meaningful principal components. Therefore, given the specific task information ($x \sim \mathcal{D}$), we define the information richness of an attention head as $I_{h}(W_{h}|x)$:
\begin{align}
    I_{h}(W_{h}|x) = \underset{t}{argmin} \left \{ \frac{\sum_{i=1}^{t}\sigma_{i}}{\sum_{j=1}^{T}\sigma_{j}} \ge \xi \right \}
    \label{eq:I_h_x}
\end{align}
where $\xi$ is a hyperparameter threshold, we set $\xi = 0.9$ in our experiments. Note that Eq.~(\ref{eq:g_W_h}) requires solving the expectation on data $x$. In practice, we approximate the solution using the Monte-Carlo algorithm:
\begin{align}
    \begin{split}
        \mathbb{E}_{x\sim \mathcal{D}} \left[ g(O_{h}(x)) \right] 
        &= \mathbb{E}_{x \sim \mathcal{D}} \left[ I_{h}(W_{h}|x) \right] \\
        &\approx \frac{1}{n} \sum_{i=1}^{n} I_{h}(W_{h}|x_{i})
    \end{split}
    \label{eq:E_I}
\end{align}
Although this operation is expensive when the amount of training data is large, we fortunately find that the stable results can be obtained using a small $n$ (e.g., $300$) in actual experiments. Besides, this metric also remains robust under diverse settings as shown in Fig.~\ref{fig:info_richness}.

\subsection{Correlation}
\label{sub_sec:correlation}
Applying a similar idea from the solution of $I_{h}$, the correlation between weights is transformed into the corresponding outputs. We define the correlation $r(W_{h}, W_{h'})$ between two heads ($h, h'$) as:
\begin{align}
    r(W_{h}, W_{h'}) = \mathbb{E}_{x \sim \mathcal{D}} \left[ r(O_{h}, O_{h'} | x) \right]
    \label{eq:define_r}
\end{align}
where the outputs $O_{h}, O_{h'} \in \mathbb{R}^{S\times D'}$. To calculate $r(O_{h}, O_{h'} | x)$, we first average $O_{h}$ over the sequence axis, i.e., $O'_{h} = \frac{1}{S} \sum_{s=1}^{S} O_{h}(s, :)$, where $O_{h}(s, :) \in \mathbb{R}^{D'}$ refers to the hidden representation of the $s$-th token in the sequence.

Next, the correlation between two heads ($h, h'$) is computed by the covariance:
\begin{align}
    r(O'_{h}, O'_{h'} | x) = \Big| \mathrm{cov}(O'_{h}(x), O'_{h'}(x)) \Big|
    \label{eq:r_x}
\end{align}
Here, we are focusing on the degree of correlation, where the correlation for strong positive and negative should be considered equally. Hence, we put the absolute operation on Eq.~(\ref{eq:r_x}). We apply the unbiased estimation of covariance:
\begin{align}
    \mathrm{cov}(O'_{h}, O'_{h'}) = \frac{\sum_{d=1}^{D'} (o_{h,d} - \bar{o}_{h}) (o_{h',d} - \bar{o}_{h'})}{D'-1}
    \label{eq:cov}
\end{align}
where $o_{h,d}$ and $o_{h',d}$ represent the $d$-th element in $O'_{h}$ and $O'_{h'}$, while $\bar{o}_{h}$ and $\bar{o}_{h'}$ indicate the average of $O'_{h}$ and $O'_{h'}$, respectively. Thus, the correlation matrix of heads is defined as $R = [r_{h, h'}]_{H \times H}$, the entry of which is $r_{h, h'} \doteq \mathbb{E}_{x\sim\mathcal{D}} \left[ r(O_{h}, O_{h'} | x ) \right]$ and $r_{h, h} = 0$. For the calculation of Eq.~(\ref{eq:define_r}), we also adopt the Monte-Carlo algorithm to approximate the solution:
\begin{align}
    \mathbb{E}_{x\sim\mathcal{D}} \left[ r(O_{h}, O_{h'} | x ) \right] \approx \frac{1}{n} \sum_{i=1}^{n} r(O_{h}, O_{h'} | x_{i} )
    \label{eq:E_r}
\end{align}
We show the head-to-head correlation heatmap in Fig.~\ref{fig:corr_st}. In addition, we report more comparisons in appendix \ref{appendix:settings_IR_CORR} to illustrate its robustness under different experimental settings.

\begin{figure*}[!t]
    \centering
    \includegraphics[width=1.0\textwidth]{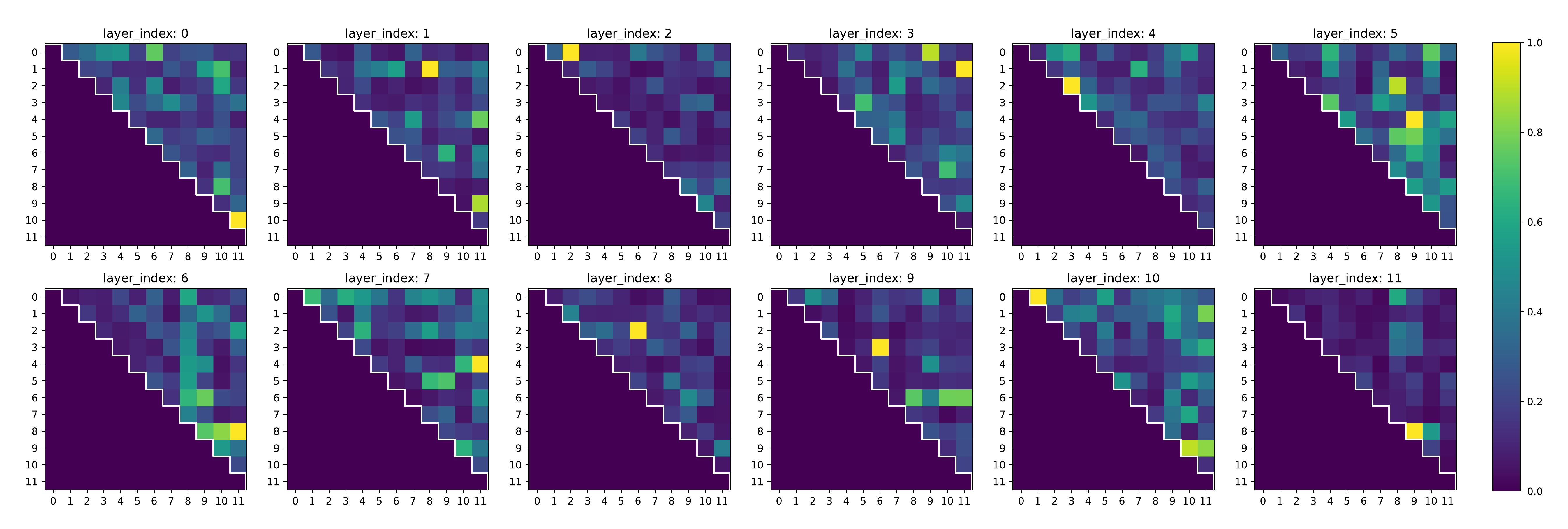}
    \caption{The head-to-head correlation ($r_{h, h'}$). In each subfigure, both $x$-axis and $y$-axis represent the index of heads, and the value of correlation is normalized. See Sec.~\ref{sub_sec:robustness} for detailed experimental settings.}
    \label{fig:corr_st}
    \vspace{-1.0em}
\end{figure*}

\subsection{Joint Optimization}
To summarize, we can obtain $I_{h}$, which measures the information richness of a head, and $R$, which indicates the correlation matrix between heads. To determine the relative importance of each head, we first model the relationship between heads as a directed fully-connected graph, as shown in Fig.~\ref{fig:overview}. In this graph, the initial probability $p_{h}^{(0)}$ per node (i.e., head) is defined as:
\begin{align}
    p_{h}^{(0)} = \frac{I_{h}}{\sum_{h'=1}^{H} I_{h'}}
    \label{eq:p_h_0}
\end{align}
Then, we define $m_{h, h'}$, the probability of moving from node $h$ to another node $h'$, as:
\begin{align}
    m_{h, h'} = \frac{r_{h, h'}}{\sum_{h''\ne h}^{H} r_{h, h''}}
    \label{eq:m_h_h'}
\end{align}
Hence, we can obtain the initial probability vector $P^{(0)} = [p_{1}^{(0)}, p_{2}^{(0)}, \cdots, p_{H}^{(0)}]^{\top}$ and the state transition probability matrix $M = [m_{h, h'}]_{H \times H}$. Given that $H$ is generally small in practice (e.g., $H=16$ in $\text{BERT}_{\text{LARGE}}$), we employ the iterative method of PageRank \cite{page1999pagerank} to search for the optimal solution:
\begin{align}
    P^{(t+1)} = d M P^{(t)} + \frac{1-d}{H}\mathbb{I}
    \label{eq:P_t+1}
\end{align}
where $d$ refers to the damping factor, and $\mathbb{I}$ is the $H$-dimensional vector with all elements of $1$. From the perspective of PageRank \cite{page1999pagerank}, when the Markov chain reaches stationary distribution, the PageRank value $p_{h}^{*}$ per node in the graph can be obtained: $P^{*} = \underset{t\to\infty}{\lim} P^{(t+1)} = [p_{1}^{*}, p_{2}^{*}, \cdots, p_{H}^{*}]^{\top}$.
Finally, we utilize $p_{h}^{*}$ to evaluate the relative importance of the $h$-th head, and then take the top-$k$ heads for fine-tuning. The detailed algorithm procedures are summarized in appendix \ref{appendix:algorithm}.

\section{Experiments}
\subsection{Setup}
\label{sub_sec:setup}
\paragraph{Datasets \& Evaluation Protocol.}
Following the previous setting \cite{houlsby2019parameterefficient}, we evaluate our method on the GLUE benchmark \cite{wang2018glue}, which consists of eight datasets (i.e., CoLA, SST-2, MPRC, QQP, STS-B, MNLI, QNLI, and RTE). See \ref{appendix:datasets} in appendix for detailed description per dataset. In addition, Matthew's (Matt.) and Spearman's (Spea.) correlation coefficient are used to test CoLA and STS-B, respectively. MRPC and QQP are measured by F1 score. As for other datasets, the accuracy (Acc.) metric is applied.

\paragraph{Full-shot Learning.} According to the vanilla evaluation procedure, we utilize the datasets library\footnote{\url{https://github.com/huggingface/datasets}} to load the complete dataset for training, and then save the best checkpoint based on the performance on validation set, and finally report the results on test set by submitting our predictions to the online evaluator\footnote{\url{https://gluebenchmark.com/}}.

\paragraph{Few-shot Learning.} Following the setting of \citet{sun2022blackbox}, we randomly select $16$ samples per class to construct $16$-shot training set $\mathcal{D}_{\text{train}}$ and validation set $\mathcal{D}_{\text{val}}$ from the original training set, respectively. In addition, the original validation set is regarded as the test set $\mathcal{D}_{\text{test}}$, where $|\mathcal{D}_{\text{train}}| = |\mathcal{D}_{\text{val}}| \ll |\mathcal{D}_{\text{test}}|$. The STS-B dataset is excluded since it is a regressive task.

\begin{table*}[!t]
    \centering
    \scalebox{0.8}{
        \begin{tabular}{l | c c c c c c c c c c}
            \toprule
                Model & QNLI & SST-2 & $\text{MNLI}_{m}$ & $\text{MNLI}_{mm}$ & CoLA & MRPC & STS-B & RTE & QQP & Avg. \\
                      & (Acc.) & (Acc.) & (Acc.) & (Acc.) & (Matt.) & (F1) & (Spea.) & (Acc.) & (F1) & \\
            \midrule
                \multicolumn{11}{c}{Full-shot Learning} \\
            \midrule
                $\text{Full-FT}^{\dagger}$ & $\underline{93.4}$ & $94.1$ & $\textbf{86.7}$ & $\textbf{86.0}$ & $59.6$ & $88.9$ & $86.6$ & $71.2$ & $71.7$ & $\underline{82.0}$ \\ % from Diff-Pruning
            \midrule
                $\text{Diff-Pruning}^{\dagger}$ & $93.3$ & $94.1$ & $\underline{86.4}$ & $\textbf{86.0}$ & $\textbf{61.1}$ & $\textbf{89.7}$ & $86.0$ & $70.6$ & $71.1$ & $\underline{82.0}$ \\
                Child-Tuning & $92.6_{\pm 0.3}$ & $\underline{94.2}_{\pm 0.5}$ & $86.1_{\pm 0.6}$ & $85.2_{\pm 0.4}$ & $59.2_{\pm 0.5}$ & $88.1_{\pm 0.8}$ & $85.3_{\pm 0.5}$ & $71.2_{\pm 0.5}$ & $71.3_{\pm 0.3}$ & $81.5$ \\
            \midrule
                $\text{Adapter}^{\dagger}$ & $90.7$ & $94.0$ & $84.9$ & $85.1$ & $59.5$ & $\underline{89.5}$ & $\underline{86.9}$ & $\underline{71.5}$ & $\underline{71.8}$ & $81.5$ \\ % 3.6% _{8-256}
                
                $\text{BitFit}^{\dagger}$ & $92.0$ & $\underline{94.2}$ & $84.5$ & $84.8$ & $59.7$ & $88.9$ & $85.5$ & $\textbf{72.0}$ & $70.5$ & $81.3$ \\
                
                LoRA & $91.2_{\pm 0.5}$ & $93.2_{\pm 0.3}$ & $84.2_{\pm 0.7}$ & $84.1_{\pm 0.5}$ & $60.2_{\pm 0.9}$ & $88.8_{\pm 0.7}$ & $85.9_{\pm 0.2}$ & $70.3_{\pm 0.3}$ & $71.0_{\pm 0.5}$ & $81.0$ \\
                
                Compacter & $91.5_{\pm 0.2}$ & $93.6_{\pm 0.4}$ & $85.3_{\pm 0.5}$ & $84.9_{\pm 0.3}$ & $58.6_{\pm 0.6}$ & $87.9_{\pm 1.0}$ & $86.6_{\pm 0.6}$ & $69.7_{\pm 0.4}$ & $\textbf{71.8}_{\pm 0.2}$ & $81.1$ \\
                
                Prefix-Tuning & $92.2_{\pm 0.5}$ & $\textbf{94.3}_{\pm 0.3}$ & $84.2_{\pm 0.3}$ & $84.0_{\pm 0.4}$ & $58.4_{\pm 0.8}$ & $88.2_{\pm 0.5}$ & $85.7_{\pm 0.3}$ & $71.3_{\pm 0.2}$ & $69.7_{\pm 0.6}$ & $80.9$ \\

            \midrule
                $\textbf{HiFi}_{\textbf{mid-top}}$ & $92.7_{\pm 0.2}$ & $93.9_{\pm 0.5}$ & $85.8_{\pm 0.3}$ & $85.5_{\pm 0.6}$ & $59.1_{\pm 0.9}$ & $88.6_{\pm 0.6}$ & $86.8_{\pm 0.2}$ & $70.8_{\pm 0.5}$ & $71.5_{\pm 0.8}$ & $81.6$ \\

                $\textbf{HiFi}_{\textbf{layer-wise}}$ & $\textbf{93.5}_{\pm 0.6}$ & $\textbf{94.3}_{\pm 0.3}$ & $85.9_{\pm 0.7}$ & $\underline{85.8}_{\pm 0.4}$ & $\underline{60.4}_{\pm 0.7}$ & $\textbf{89.7}_{\pm 0.4}$ & $\textbf{87.2}_{\pm 0.3}$ & $\underline{71.5}_{\pm 0.2}$ & $\textbf{72.0}_{\pm 0.4}$ & $\textbf{82.3}$ \\
                
            \midrule
                \multicolumn{11}{c}{Few-shot Learning} \\
            \midrule
                Full-FT & $67.9_{\pm 3.8}$ & $72.4_{\pm 6.7}$ & $44.1_{\pm 5.4}$ & $45.1_{\pm 5.6}$ & $\textbf{28.5}_{\pm 5.5}$ & $73.3_{\pm 6.3}$ & - & $54.5_{\pm 4.5}$ & $59.5_{\pm 4.8}$ & $55.7$ \\
            \midrule
                Diff-Pruning & $63.1_{\pm 1.3}$ & $74.0_{\pm 5.4}$ & $42.5_{\pm 6.1}$ & $40.6_{\pm 5.2}$ & $21.1_{\pm 8.9}$ & $72.7_{\pm 2.8}$ & - & $53.5_{\pm 3.1}$ & $57.8_{\pm 4.5}$ & $53.2$ \\
                Child-Tuning & $65.8_{\pm 2.2}$ & $76.1_{\pm 4.6}$ & $40.7_{\pm 4.3}$ & $41.4_{\pm 3.7}$ & $24.7_{\pm 7.5}$ & $73.1_{\pm 3.9}$ & - & $52.5_{\pm 4.2}$ & $58.3_{\pm 3.8}$ & $54.1$ \\
            \midrule
                Adapter & $67.2_{\pm 3.3}$ & $78.3_{\pm 4.6}$ & $42.2_{\pm 5.0}$ & $44.5_{\pm 5.7}$ & $26.6_{\pm 4.2}$ & $73.0_{\pm 7.6}$ & - & $55.0_{\pm 2.2}$ & $59.2_{\pm 2.9}$ & $55.8$ \\

                BitFit & $\textbf{70.3}_{\pm 2.1}$ & $78.9_{\pm 7.4}$ & $43.5_{\pm 3.4}$ & $42.9_{\pm 4.4}$ & $23.8_{\pm 9.2}$ & $72.4_{\pm 4.8}$ & - & $54.7_{\pm 1.1}$ & $\underline{61.3}_{\pm 4.3}$ & $56.0$ \\

                LoRA & $68.8_{\pm 1.9}$ & $\textbf{80.3}_{\pm 5.2}$ & $41.3_{\pm 2.7}$ & $42.9_{\pm 3.0}$ & $\underline{27.1}_{\pm 7.2}$ & $\underline{75.8}_{\pm 5.9}$ & - & $55.2_{\pm 1.9}$ & $60.5_{\pm 6.1}$ & $\underline{56.5}$ \\
                
                Compacter & $\underline{69.6}_{\pm 1.8}$ & $76.6_{\pm 9.5}$ & $43.4_{\pm 6.0}$ & $45.4_{\pm 6.9}$ & $25.9_{\pm 8.5}$ & $73.5_{\pm 7.2}$ & - & $52.4_{\pm 3.4}$ & $60.8_{\pm 4.6}$ & $56.0$ \\
                
                Prefix-Tuning & $68.3_{\pm 3.6}$ & $\underline{79.2}_{\pm 1.9}$ & $43.3_{\pm 4.5}$ & $\underline{45.7}_{\pm 4.8}$ & $24.8_{\pm 3.3}$ & $72.4_{\pm 9.3}$ & - & $54.4_{\pm 2.5}$ & $60.4_{\pm 2.3}$ & $56.1$ \\
            \midrule
                $\textbf{HiFi}_{\textbf{mid-top}}$ & $67.7_{\pm 1.6}$ & $76.2_{\pm 2.4}$ & $\underline{44.8}_{\pm 4.4}$ & $44.7_{\pm 5.3}$ & $26.8_{\pm 6.5}$ & $75.2_{\pm 5.2}$ & - & $\underline{55.4}_{\pm 3.8}$ & $59.7_{\pm 3.9}$ & $56.3$ \\
                
                $\textbf{HiFi}_{\textbf{layer-wise}}$ & $68.5_{\pm 2.7}$ & $76.6_{\pm 3.5}$ & $\textbf{45.9}_{\pm 5.7}$ & $\textbf{45.8}_{\pm 6.3}$ & $\textbf{28.5}_{\pm 5.2}$ & $\textbf{76.3}_{\pm 3.5}$ & - & $\textbf{55.5}_{\pm 3.5}$ & $\textbf{61.8}_{\pm 4.1}$ & $\textbf{57.4}$ \\
            \bottomrule
        \end{tabular}
    }
    \caption{The performance on the GLUE benchmark. The results are averaged from three seeds in the full-shot learning\protect\footnotemark, while five seeds are used in the few-shot learning to produce solid results. The subscript is the standard deviation. \textbf{Bold} and \underline{underline} indicate the first and second best results in the corresponding regime. $\dagger$ refers to the results directly from their original paper, in which Full-FT is derived from \citet{guo2021parameterefficient}.}
    \label{tab:glue}
    \vspace{-1.0em}
\end{table*}
\footnotetext{The average score of $\dagger$ differs slightly from the original paper due to the different averaging method, e.g., BitFit first averages $\text{MNLI}_{m}$ and $\text{MNLI}_{mm}$, and then calculates the overall average score. Here, we directly calculate the average score of all datasets on GLUE, in a unified manner.}

\paragraph{Baselines.} To make a comprehensive comparison, we first select full fine-tuning (Full-FT) as a strong baseline, then choose five structured methods (i.e., Adapter \cite{houlsby2019parameterefficient}, Compactor \cite{mahabadi2021compacter}, Prefix-Tuning \cite{li2021prefixtuning}, LoRA \cite{hu2022lora} and BitFit \cite{benzaken2022bitfit}), and two non-structured methods (i.e., Diff-Pruning \cite{guo2021parameterefficient}, and Child-Tuning \cite{xu2021raise}) as the other baselines. See \ref{appendix:baselines} in appendix for more description of each baseline.

\paragraph{Models \& Implementation.}
Given that, in pre-training, the lower layers of PLMs learn general semantic features, which might be universal across tasks \cite{houlsby2019parameterefficient, ruckle2021adapterdrop}. Therefore, two models are proposed here: $\textbf{HiFi}_{\textbf{layer-wise}}$ fine-tunes the selected top-$k$ heads at each layer; $\textbf{HiFi}_{\textbf{mid-top}}$ only updates the layers from middle to top, while keeping the layers from bottom to middle frozen. $k$ is set to $3$ by default and its implications are further explored in Sec.~\ref{sub_sec:influence}. We leverage $\text{BERT}_{\text{LARGE}}$ as backbone and implement our models by Huggingface's Transformers library \cite{wolf2020transformers}. The off-the-shelf Adapter-Transformers library\footnote{\url{https://github.com/Adapter-Hub/adapter-transformers}} \cite{pfeiffer2020adapterhub} is utilized to perform the baselines. See \ref{appendix:settings} in appendix for more detailed experimental configurations.

\subsection{Results}
In Tab.~\ref{tab:glue}, we show the results of our models and baselines on the GLUE benchmark. From a wide range of comparisons, we can obtain that:
(\romannumeral 1) Under the full-shot setting, our proposed $\textbf{HiFi}_{\textbf{layer-wise}}$ obtains the average of $82.3$, which outperforms Full-FT ($82.0$). Meanwhile, $\textbf{HiFi}_{\textbf{mid-top}}$ also achieves satisfactory performance ($81.6$), in the case of tuning only half of the trainable parameters compared to $\textbf{HiFi}_{\textbf{layer-wise}}$.
Besides, the non-structured methods achieve better performance than the prior structured methods and are closer to the results of Full-FT, but none of them exceed Full-FT on average.
(\romannumeral 2) In the few-shot setting, the structured methods have significant advantages over Full-FT and non-structured methods. Our models (including the previous structured methods) achieve higher average score than Full-FT, while the non-structured methods are substantially lower than Full-FT\footnote{For an example of Diff-Pruning, the reason is that it probably fails to optimize the perturbation variable for each original parameter with such small training samples.}.

In short, our proposed HiFi bridges the gap between structured and non-structured methods in the full-shot setting, while maintaining a significant lead in the few-shot setting. On one hand, as we have discussed before, those heads we selected are vital for downstream tasks and could greatly influence model decisions; on the other hand, it acts as a regularizer to fine-tune only a subset of the full parameters, which boosts the generalization capacity. We further verify this hypothesis from the perspective of loss landscape in Sec.~\ref{sub_sec:perspective}.

\begin{figure*}[!t]
    \centering
    \includegraphics[width=\linewidth]{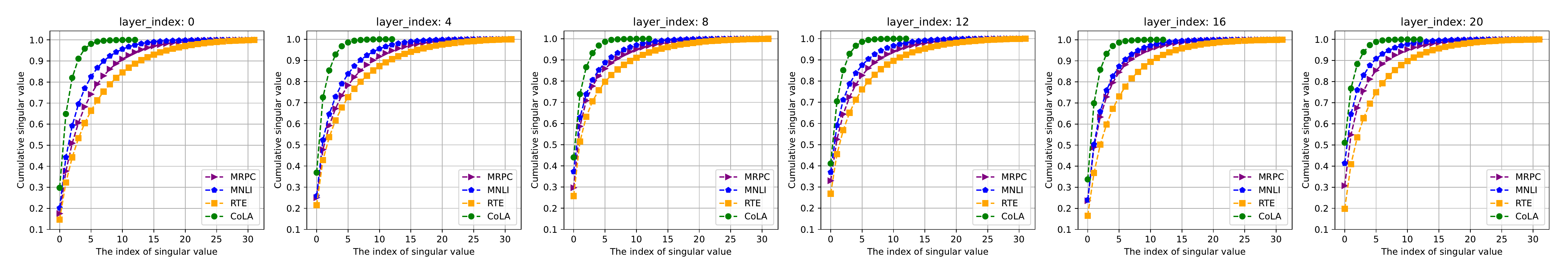}
    \caption{The effect of cumulative singular value on various datasets. We randomly select the output of a head in different layers for illustration.}
    \label{fig:effect_xi}
\end{figure*}

\begin{table*}[!t]
    \centering
    \scalebox{0.85}{
        \begin{tabular}{l | c c c c c c c c c c}
            \toprule
                Model & QNLI & SST-2 & $\text{MNLI}_{m}$ & $\text{MNLI}_{mm}$ & CoLA & MRPC & STS-B & RTE & QQP & Avg. \\
            \midrule
                \textbf{HiFi}                  & 93.2 & 94.5 & 85.9 & 85.8 & 61.5 & 92.6 & 88.1 & 73.3 & 87.8 & 84.7 \\
                \textbf{HiFi (w/o corr)}       & 92.2 & 93.2 & 85.4 & 85.1 & 61.1 & 92.0 & 88.0 & 72.8 & 87.2 & 84.1 \\
                \textbf{HiFi (w/o corr + inv)} & 91.8 & 93.5 & 84.8 & 84.6 & 60.8 & 89.1 & 87.1 & 72.0 & 86.9 & 83.4 \\
                \textbf{HiFi (w/o info)}       & 92.6 & 93.4 & 84.4 & 84.2 & 61.1 & 91.4 & 87.8 & 72.2 & 87.5 & 83.8 \\
                \textbf{HiFi (page inv)}       & 92.4 & 93.6 & 84.3 & 84.5 & 60.4 & 91.2 & 87.6 & 72.8 & 86.8 & 83.7 \\
                \textbf{HiFi (random)}         & 90.2 & 92.4 & 83.5 & 83.1 & 59.3 & 88.5 & 86.5 & 70.6 & 86.0 & 82.4 \\
            \bottomrule
        \end{tabular}
    }
    \caption{Ablation experiments.}
    \label{tab:ablation}
\end{table*}

\section{Analysis\footnote{In this section, we conduct the experiments based on $\text{HiFi}_{\text{layer-wise}}$, unless otherwise specified.}}
\subsection{Robustness of Metrics}
\label{sub_sec:robustness}
In this subsection, we study the following question: whether we designed metrics ($I_{h}, r_{h, h'}$) are robust enough under diverse settings? To this end, we take $\text{BERT}_{\text{BASE}}$ ($12$ layers) as an example to conduct experiments on MRPC\footnote{In fact, the consistent observation emerges on $\text{BERT}_{\text{LARGE}}$ and different datasets.}, in terms of batch size, sequence length, sample size (i.e., $n$) and learning rate. Specifically, \textbf{ST} is a standard baseline, where the batch size is $32$, the learning rate is $2e^{-5}$, the maximum sequence length is $128$ and the sample size is $300$. Compared to \textbf{ST}, \textbf{BS} reduces the batch size to $16$, \textbf{LR} increases the learning rate to $5e^{-5}$, \textbf{SL} reduces the maximum sequence length to $64$, and \textbf{SS} increases the number of samples to $1000$.

The results are shown in Fig.~\ref{fig:info_richness}, \ref{fig:corr_st} (partially in appendix \ref{appendix:settings_IR_CORR}), we can draw the following findings:
(\romannumeral 1) For the information richness ($I_{h}$), even if the experimental settings are diverse, $I_{h}$ almost remains consistent (e.g., in $2, 4, 6$ layers). Note that $I_{h}$ will be normalized by Eq.~(\ref{eq:p_h_0}), the absolute ranking of $I_{h}$ is inessential (e.g., in $10, 11$ layers, the curves are slightly moved up and down), as long as the relative ranking across heads remains stable.
(\romannumeral 2) For the correlation ($r_{h, h'}$), the head-to-head correlation in the heatmap is varying across layers, which means that this metric has good distinguishability. In addition, although the correlation region varies slightly in diverse settings, the strongly correlated heads are almost unchanged\footnote{Compare Fig.~\ref{fig:corr_st} with Fig.~\ref{fig:corr_bs}-\ref{fig:corr_size} in appendix, where each figure is produced in a setting.}.
This demonstrates that $I_{h}$ and $r_{h, h'}$ have good robustness and capture some essential characteristics of heads.

\subsection{Effectiveness of Methods}
\label{sub_sec:effectiveness}
Here, we focus on investigating the possible questions:
Q1: Does the correlation ($r_{h, h'}$) between heads really matter?
Q2: Are the higher information richness ($I_{h}$) of heads more important for the model?
Q3: Is it enough to only take the correlation ($r_{h, h'}$) into consideration, while ignoring the information richness ($I_{h}$)?
Q4: Does PageRank algorithm really work?

To answer the above questions, we design the following experiments for verification:
(\romannumeral 1) For Q1, we exclude the correlation ($r_{h, h'}$) compared to the baseline, and update $k$ heads corresponding the top information richness. This experiment denotes \textbf{HiFi (w/o corr)}.
(\romannumeral 2) For Q2, in contrast to Q1, we merely update $k$ heads with the lowest information richness, without taking $r_{h,h'}$ into account. This experiment denotes \textbf{HiFi (w/o corr + inv)}.
(\romannumeral 3) For Q3, the information richness is not included and only $k$ heads with the strongest correlation are updated. This experiment denotes \textbf{HiFi (w/o info)}.
(\romannumeral 4) For Q4, compared to the baseline, we inversely select $k$ heads with the lowest PageRank value. This experiment denotes \textbf{HiFi (page inv)}.
In addition, to establish the effectiveness of our selected heads, we also compare with a baseline that randomly tunes $k$ heads in each layer. This experiment denotes \textbf{HiFi (random)}.
Here, the experiments are conducted on the validation sets of GLUE, and the comparison results are shown in Tab.~\ref{tab:ablation}. We observe that the optimal performance is only obtained when the information richness ($I_{h}$) and correlation ($r_{h, h'}$) are jointly addressed by PageRank algorithm.

\begin{figure}[!t]
    \centering
    \subfigure {
        \includegraphics[width=0.22\textwidth]{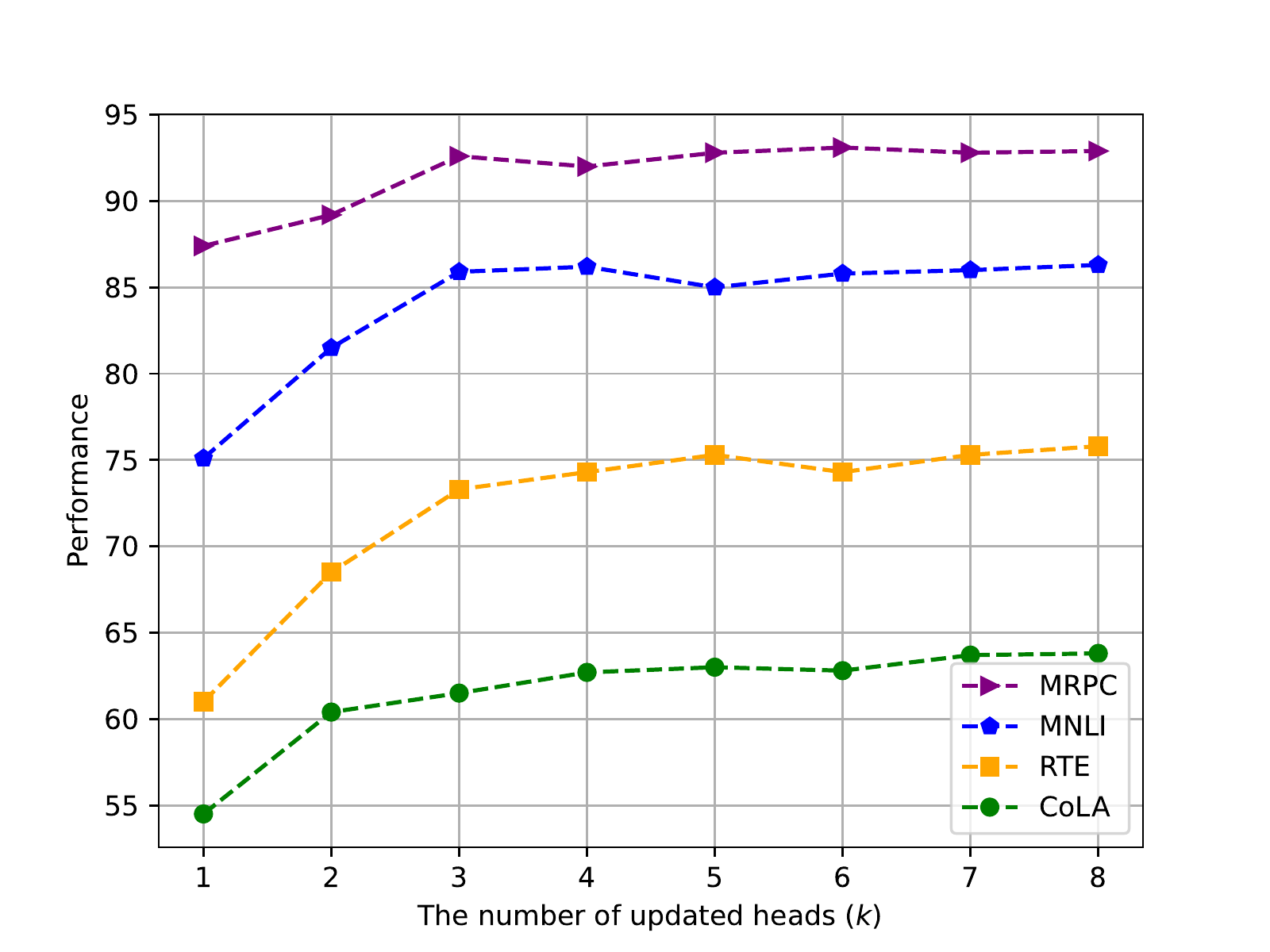}
    }
    \subfigure {
        \includegraphics[width=0.22\textwidth]{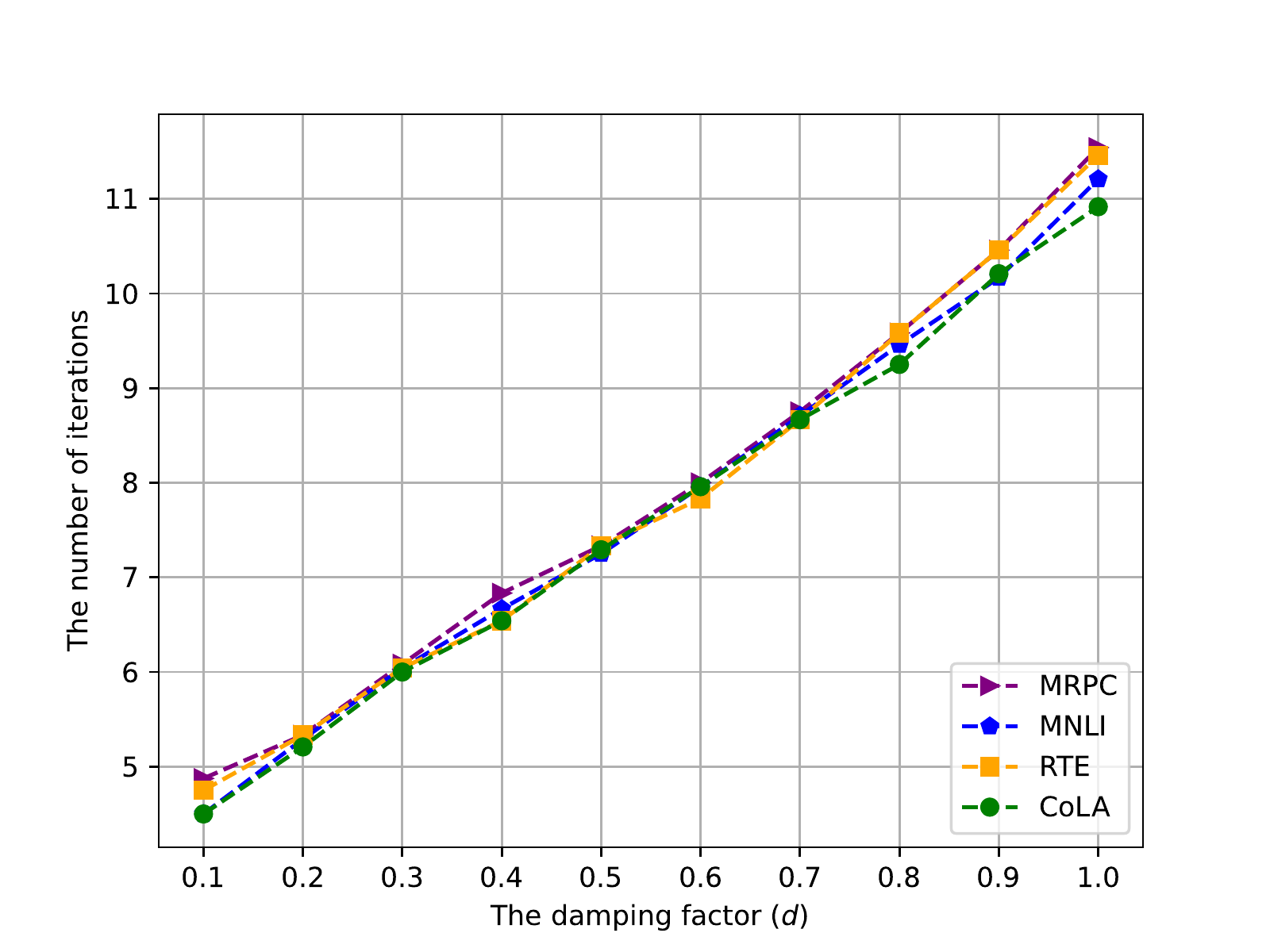}
    }
    \caption{The effect of the number of selected heads $k$ (Left) and the damping factor $d$ (Right).}
    \label{fig:effect_head_damping}
\end{figure}

\subsection{Influence of Hyperparameters}
\label{sub_sec:influence}
In this subsection, we probe the following hyperparameters: the number of selected heads ($k$), the proportion of the top-$t$ singular values ($\xi$) and the damping factor ($d$), based on MPRC, CoLA, RTE and $\text{MNLI}$, respectively.
(\romannumeral 1) $k$ is a key factor to control the ratio of trainable parameters over the all parameters. In Fig.~\ref{fig:effect_head_damping}, the optimal value of $k$ is around $3$ and the corresponding ratio is $4.2\%$. If the mid-top strategy is adopted, the ratio can be further reduced to $2.1\%$\footnote{In fact, this ratio can be further decreased as the model scale increases because most parameters of Transformer block are concentrated in the feedforward ($66.7\%$), while MHA is only $33.3\%$, in which we fine-tune several heads.}.
(\romannumeral 2) The curves have greater curvature when the $y$-axis reaches around $0.9$, and the growth subsequently becomes slow as shown in Fig.~\ref{fig:effect_xi}, which indicates that the top-$t$ principal components are highly informative. Therefore, $\xi = 0.9$ is a good boundary.
(\romannumeral 3) In Fig.~\ref{fig:effect_head_damping}, the number of iterations grows as $d$ increases. Nevertheless, we find that the PageRank algorithm is fast (less than $0.4$ second) in practical experiments. More analysis about efficiency in appendix \ref{appendix:efficiency}. Besides, the convergence consistency of PageRank \cite{page1999pagerank} also guarantees the stability of our results, once $P^{(0)}$ and $M$ had been obtained.

\begin{figure}[!t]
    \centering
    \includegraphics[width=0.45\textwidth]{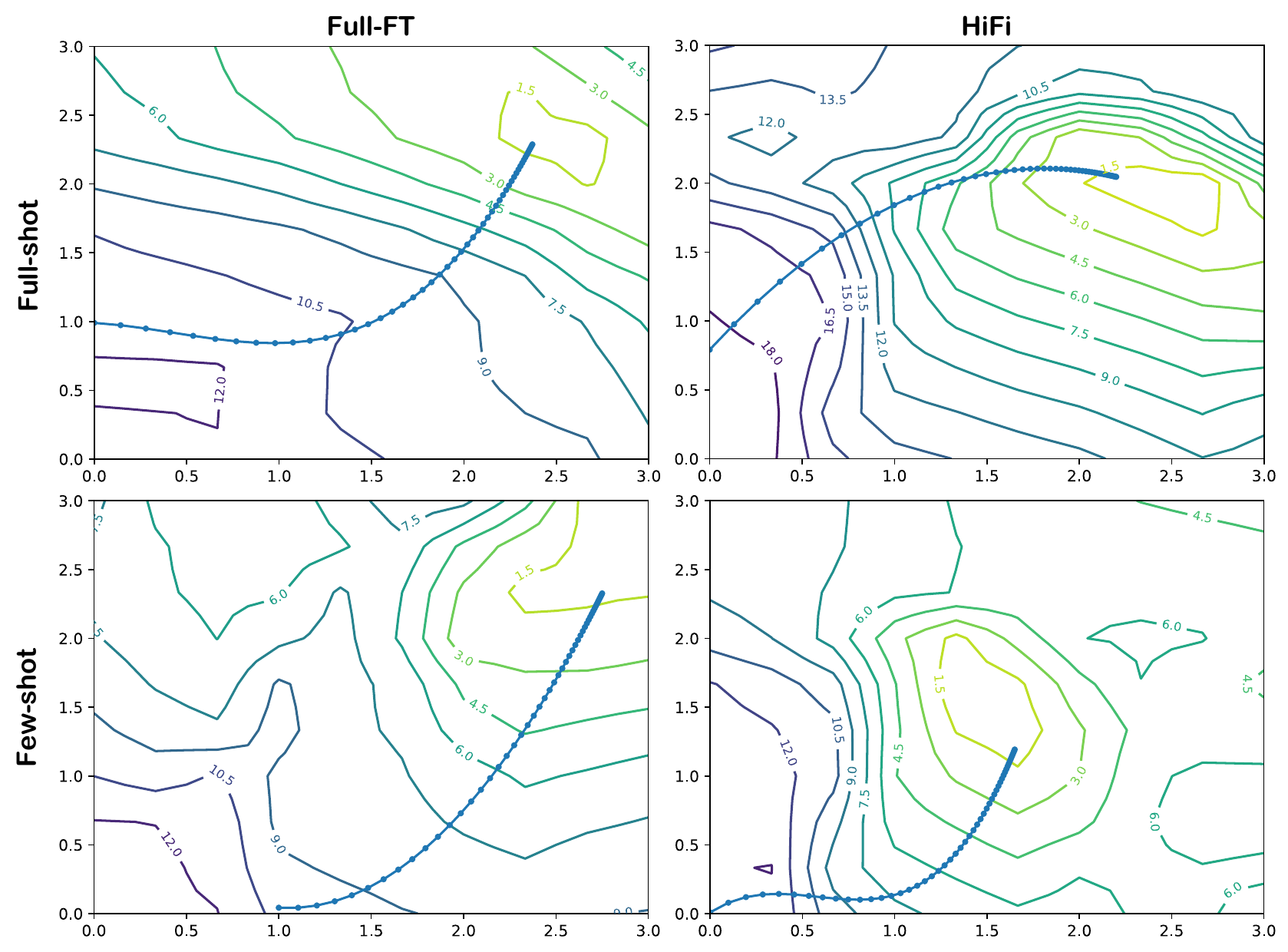}
    \caption{Visualization of loss contours and training trajectories of Full-FT (first column) and HiFi (second column), in both full-shot and few-shot scenarios.}
    \label{fig:vis_loss}
\end{figure}

\subsection{Perspective of Loss Landscape}
\label{sub_sec:perspective}
To further understand the effectiveness of HiFi compared to Full-FT, we visualize the training trajectories along with loss landscapes through tools \cite{li2018visualizing, hao2019visualizing} on MRPC. As shown in Fig.~\ref{fig:vis_loss}, the loss contour of HiFi is smoother/flatter than Full-FT in both full-shot and few-shot settings. A smoother/flatter loss surface is often believed to indicate enhanced standard and robust generalization \cite{he2019asymmetric, wu2020adversarial}, where an intuitive understanding is that if optimum basin is flat enough, it is possible to avoid the model jumping out the well-generalized region. Therefore, this provides a possible perspective to explain that HiFi has a more robust generalization capability than Full-FT on downstream tasks.

\section{Conclusion}
In this paper, we propose a novel PEFT method HiFi by modeling the relationship between attention heads as a graph and then using the PageRank to determine the relative significant heads for fine-tuning. HiFi obtains state-of-the-art performance over the prior baselines on GLUE benchmark, in both full-shot and few-shot scenarios.

\section*{Acknowledgment}
This project is supported by the Young Scientists Fund of the National Natural Science Foundation of China (Grant No. 62006201).

\section*{Limitations}
In this work, we design a parameter-efficient fine-tuning method and demonstrate its effectiveness through extensive experiments. However, there are some limitations:
(\romannumeral 1) It may be suboptimal to select the same number of heads ($k$) in each layer. In future work, we will continue to explore how to design better selection strategies (e.g., an intuitive idea is that $k$ should be layer-specific).
(\romannumeral 2) So far, the models and datasets we tested all belong to NLP. Due to time and resource limitations, we do not evaluate the related models and datasets (e.g., ViT, CIFAR) in CV.

\section*{Ethics Statement}
Our proposed HiFi can effectively reduce the resource consumption of PLMs during the fine-tuning phase, which helps to decrease carbon emissions, thus making the large-scale models more environmentally friendly and sustainable. In addition, all the models and datasets used in our experiments are publicly available and have not been reported to carry social bias against any sensitive attributes, and the proposed approach would not explicitly introduce new negative societal impacts.

% Entries for the entire Anthology, followed by custom entries
\bibliography{anthology,custom}
\bibliographystyle{acl_natbib}

\appendix
\clearpage

\section{Appendix}

\subsection{More Description about Datasets}
\label{appendix:datasets}
The GLUE benchmark covers various types of tasks and can be divided into three folds in general (detailed statistics are shown in Tab.~\ref{tab:datasets}):
\begin{itemize}
    \item \textbf{Single sentence classification task.} CoLA (The Corpus of Linguistic Acceptability) consists of books and journals from language theory, where each sentence is marked for grammaticality. There are two labels, $0$ and $1$, where $0$ indicates ungrammatical and $1$ indicates grammatical. SST-2 (The Stanford Sentiment Treebank) contains human annotations of sentences from movie reviews and their emotions. This task is to determine the emotion of a given sentence: positive or negative.
    \item \textbf{Sentence pair paraphrase task.} MRPC (The Microsoft Research Paraphrase Corpus) is a paraphrase identification task, where the corpus automatically extracts the sentence pairs from online news sources, and manually annotates whether the sentence pairs are semantically equivalent. QQP (The Quora Question Pairs), a paraphrase identification task, is a corpus of question pairs from the website Quora. The target is also to determine whether a pair of sentences is semantically equivalent. STS-B (The Semantic Textual Similarity Benchmark) is a collection of sentence pairs extracted from news headlines, video titles, image titles and natural language inference data, and each pair is annotated by humans. This task can be regarded as a fine-grained five-way paraphrase identification task.
    \item \textbf{Natural language inference task.} MNLI (The Multi-Genre Natural Language Inference Corpus), the natural language inference task, is a collection of text implication annotation for sentence pairs through crowdsourcing. Given the premise and hypothesis, the aim is to predict whether the premise contains the assumption (entailment), contradicts the assumption (contradiction) or neither (neutral). QNLI (Question-answering NLI), a natural language inference task, is converted from another dataset (The Stanford Question Answering Dataset, SQuAD $1.0$). QNLI are obtained by combining each sentence in question and context, and filtering out the sentence pair with low lexical overlap. RTE (The Recognizing Textual Entailment) is also a natural language inference task, where the samples are from news and Wikipedia. The target is to judge whether a given sentence pair is entailment or not.
\end{itemize}

\subsection{More Description about Baselines}
\label{appendix:baselines}
We give a brief introduction to baselines used in this paper:
\begin{itemize}
    \item \textbf{Full Fine-Tuning (Full-FT)} is the most prevalent fine-tuning paradigm at present. When obtaining the pre-trained model on a large-scale corpus, full parameters are updated on each specific downstream task.
    \item \textbf{Adapter} \cite{houlsby2019parameterefficient} reduces the number of trainable parameters by updating merely the additional MLPs (with bottleneck architecture), which are inserted in each layer of model.
    \item \textbf{Diff-Pruning} \cite{guo2021parameterefficient} learns a perturbation variable $\Delta \theta$ for each parameter $\theta$ in PLMs, i.e., $\theta := \theta + \Delta \theta$, where the $L_{0}$-norm constraint is imposed on the variable set $\{\Delta \theta\}$ to ensure that this set is as sparse as possible.
    \item \textbf{Child-Tuning} \cite{xu2021raise} identifies a sub-network from the original model by calculating the parametric gradient of the Fisher Information (FIM), and then the parameters of the sub-network are updated during the fine-tuning process.
    \item \textbf{Compacter} \cite{mahabadi2021compacter} can be regarded as a variant of Adapter, which further reduces the number of training parameters by introducing Kronecker product and shared weights.
    \item \textbf{Prefix-Tuning} \cite{li2021prefixtuning} concatenates the additional trainable parameters with the $K$ and $V$ in MHA. These introduced parameters are treated as continuous prompts from the perspective of prompt learning.
    \item \textbf{LoRA} \cite{hu2022lora}, from the view of low-rank decomposition, inserts two trainable low-rank matrices into the weight matrices $W^{Q}$ and $W^{K}$ per layer in a parallel manner, so as to reduce the proportion of trainable parameters and avoid inference delay.
    \item \textbf{BitFit} \cite{benzaken2022bitfit} is an extremely simple structured fine-tuning method. Only the bias terms of model are fine-tuned, leaving the rest frozen.
\end{itemize}

\subsection{More Details about Settings \& Environments}
\label{appendix:settings}
In the full-shot learning, for a range of baselines, we first refer to the hyperparameters settings in their original paper, and then perform the grid search for the batch size and learning rate. The optimal hyperparameter combinations are shown in Tab.~\ref{tab:settings}.
Under the few-shot learning, we directly use the size of the training set as the batch size and adopt the optimal learning rate in the full-shot learning. In addition, the sub-network structure of Child-Tuning is consistent with that in the full-shot setting to avoid too few samples to identify the sub-network effectively. Similarly, we perform the selection process of heads only once and keep them unchanged in the full/few-shot settings. Finally, we conduct the experiments on three NVIDIA GeForce RTX 3090 ($24$G).

\subsection{More Experiments about Correlation}
\label{appendix:settings_IR_CORR}
\begin{figure*}[!h]
    \centering
    \includegraphics[width=0.8\textwidth]{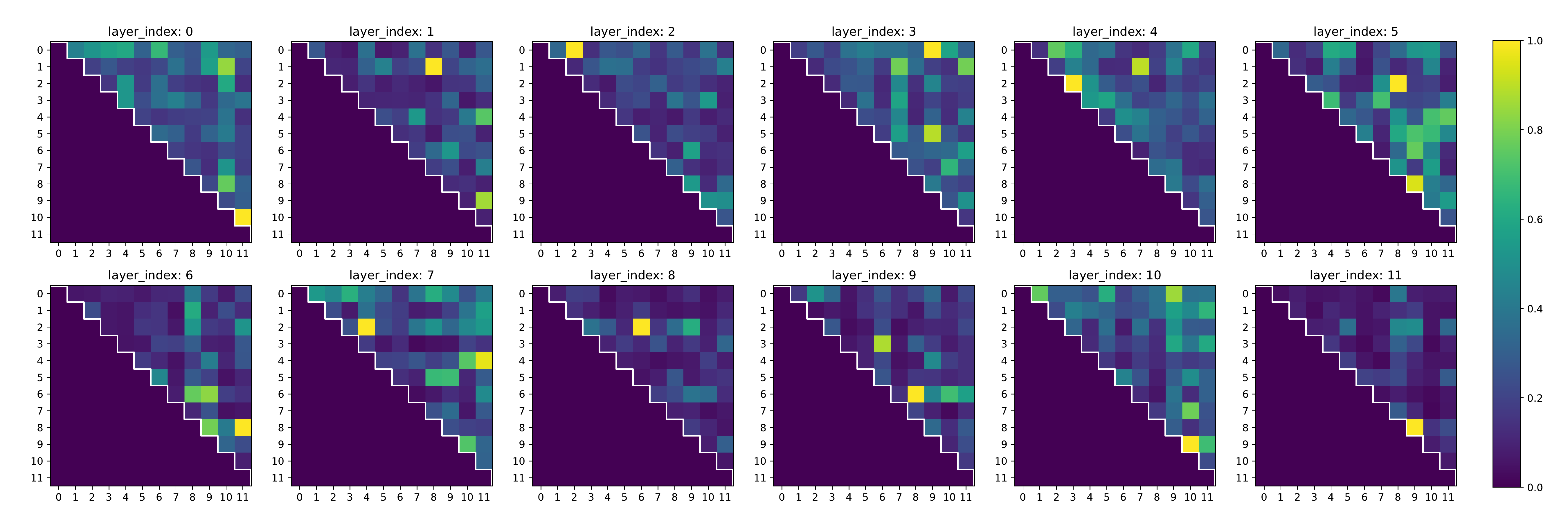}
    \caption{The effect of input batch size for the correlation between heads.}
    \label{fig:corr_bs}
\end{figure*}

\begin{figure*}[!h]
    \centering
    \includegraphics[width=0.8\textwidth]{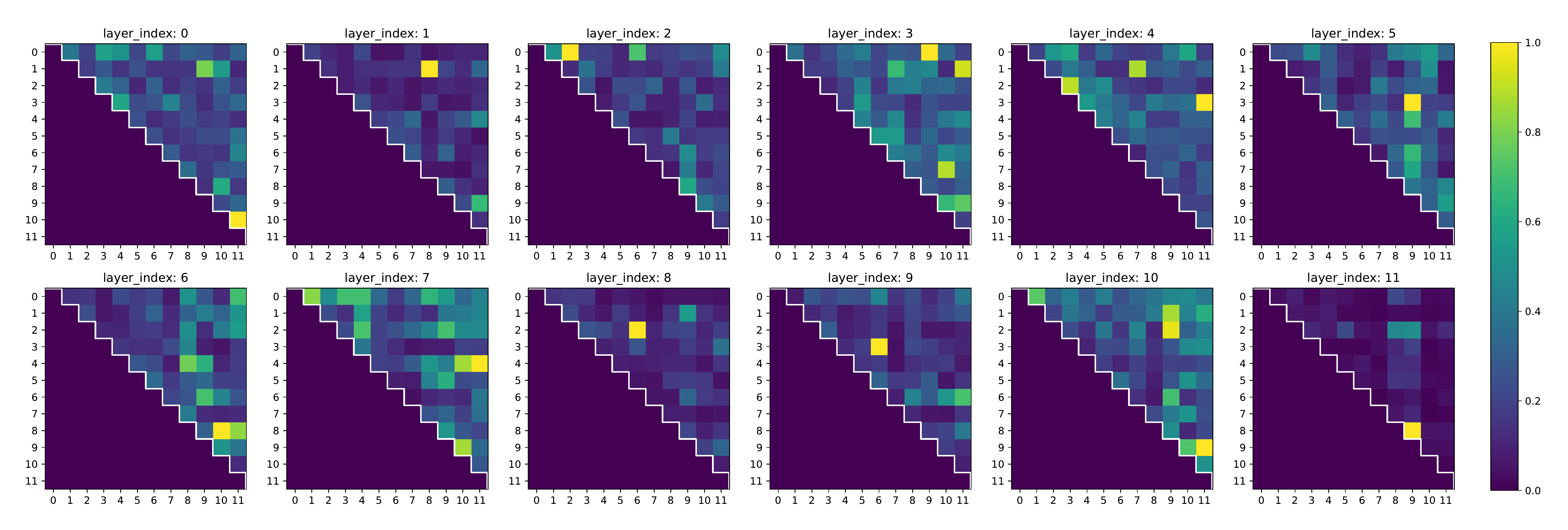}
    \caption{The effect of learning rate for the correlation between heads.}
    \label{fig:corr_lr}
\end{figure*}

\begin{figure*}[!h]
    \centering
    \includegraphics[width=0.8\textwidth]{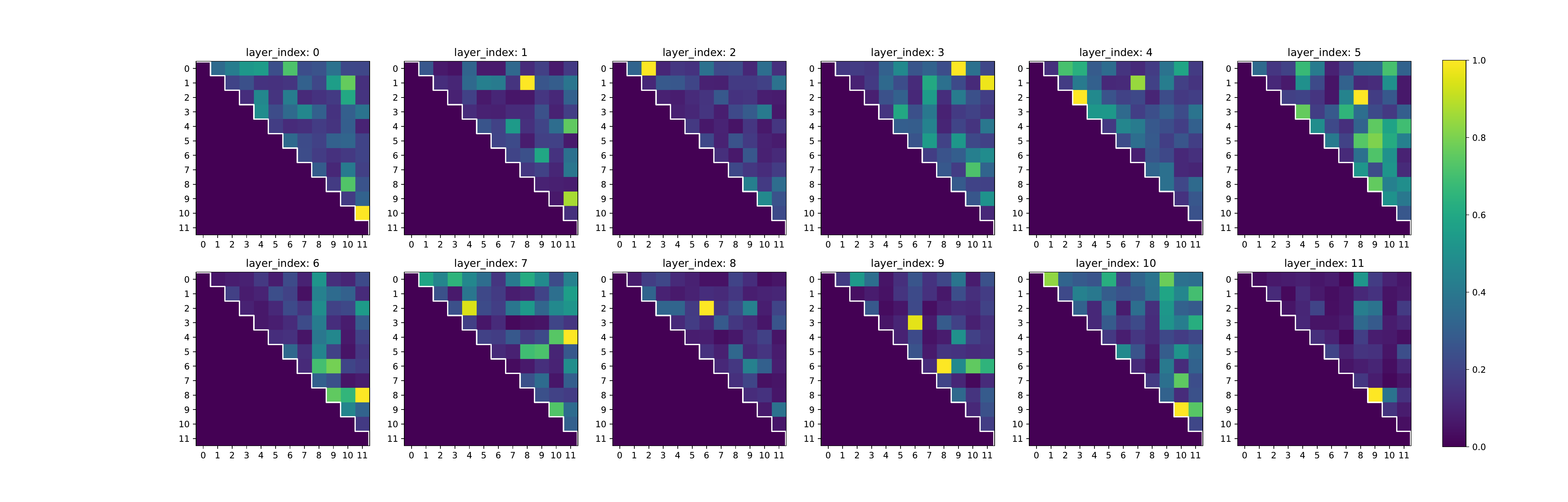}
    \caption{The effect of input sequence length for the correlation between heads.}
    \label{fig:corr_len}
\end{figure*}

\begin{figure*}[!h]
    \centering
    \includegraphics[width=0.8\textwidth]{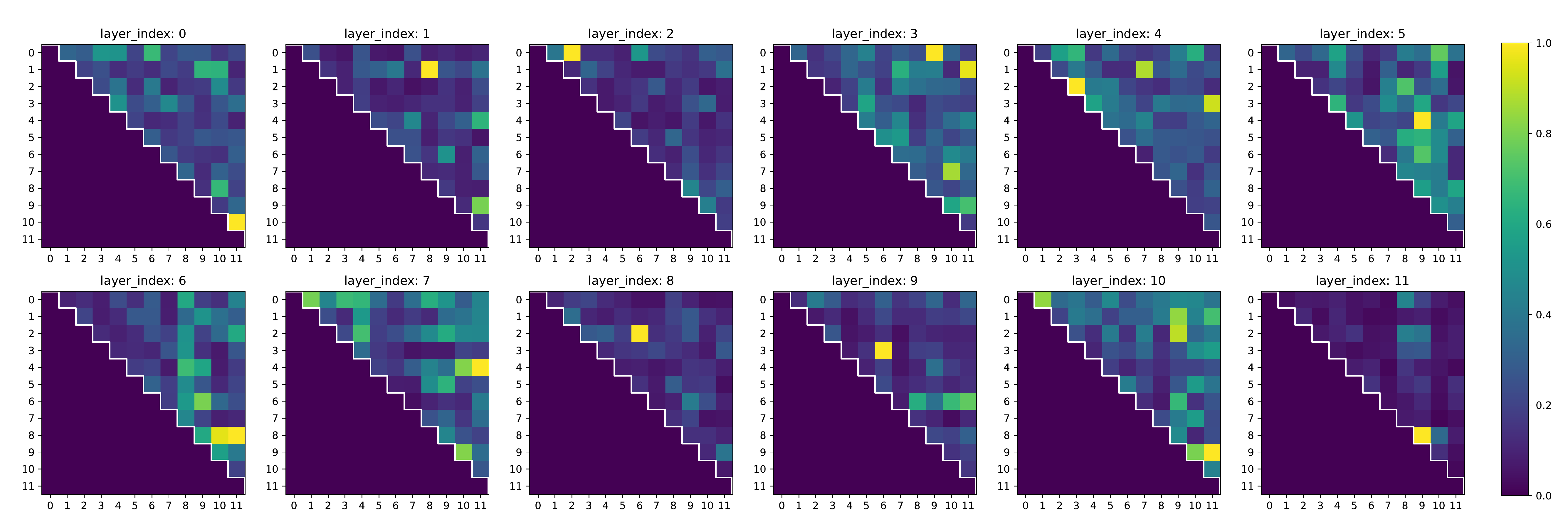}
    \caption{The effect of sample size for the correlation between heads.}
    \label{fig:corr_size}
\end{figure*}

To verify the robustness of the head-to-head correlation, we here conduct more experiments on MRPC based on $\text{BERT}_{\text{BASE}}$. The experimental results are shown in Fig.~\ref{fig:corr_bs}, \ref{fig:corr_lr}, \ref{fig:corr_len}, and \ref{fig:corr_size}, under the settings of \textbf{BS}, \textbf{LR}, \textbf{SL}, and \textbf{SS}, respectively.

\subsection{More Detailed Algorithm Procedures}
\label{appendix:algorithm}
\begin{algorithm}[!h]
    \caption{Solve for Information Richness and Correlation}
    \textbf{Input}: Training data $\mathcal{D} = \{(x_{i}, y_{i})\}_{i=1}^{N}$, the PLM $f_{\theta}(\cdot)$ and $\theta$ represents the all parameters. The number of samples is $n$. \\
    \textbf{Output}: The information richness $I_{h}$ and correlation $r_{h, h'}$.
    
    \begin{algorithmic}[1] %[1] enables line numbers
        \FOR {$1 \le i \le n$}
            \STATE Randomly sample the labelled data pair $(x_{i}, y_{i}) \sim \mathcal{D}$
            \STATE Obtain the output $O_{h}(x_{i})$ of the $h$-th head for each layer, based on $f_{\theta}(x_{i})$.
            \STATE Compute the $I_{h}(\cdot | x_{i})$.
            \STATE Compute the $r(\cdot, \cdot | x_{i})$.
        \ENDFOR
        \STATE Compute the average of $I_{h}$ and $r_{h, h'}$.
        \STATE \textbf{return} $I_{h}, r_{h, h'}$, $\forall~h, h' \in \{1, 2, \cdots, H\}$.
    \end{algorithmic}
    \label{alg:info_corr}
\end{algorithm}

\begin{algorithm}[!h]
    \caption{Joint Optimization with PageRank}
    \textbf{Input}: The information richness $I_{h}$ and correlation $r_{h, h'}$, where $h, h' \in \{1, 2, \cdots, H\}$. The upper bound of the error is $\epsilon$, and the number of updated heads is $k$. \\
    \textbf{Output}: The updated indicator $\delta_{h}$ per head.
    
    \begin{algorithmic}[1] %[1] enables line numbers
        \STATE Compute the initial node probability $p_{h}^{(0)}$ by Eq.~(\ref{eq:p_h_0}) per head.
        \STATE Compute the initial move probability from $h$ to $h'$ by Eq.~(\ref{eq:m_h_h'}).
        \STATE Obtain the initial probability matrix $P^{(0)}$ and the state transition probability matrix $M$.
        \STATE Compute the first time probability $P^{(1)}$ by Eq.~(\ref{eq:P_t+1}).
        \WHILE {$\parallel P^{(1)} - P^{(0)}\parallel > \epsilon$}
            \STATE $P^{(0)} := P^{(1)}$
            \STATE $P^{(1)} := d M P^{(1)} + \frac{1-d}{H}\mathbb{I}$
        \ENDWHILE
        \STATE Obtain the PageRank value $p_{h}^{*}$ per head.
        \STATE Let $\delta_{h} = 1$ if $h \in Topk\{p_{h}^{*}\}$, otherwise $0$.
        \STATE \textbf{return} $\delta_{h}$, $\forall~h \in \{1, 2, \cdots, H\}$.
    \end{algorithmic}
    \label{alg:pagerank}
\end{algorithm}

\begin{algorithm}[!h]
  \caption{Parameter-Efficient Fine-tuning with HiFi}
    \textbf{Input}: Training data $\mathcal{D} = \{(x_{i}, y_{i})\}_{i=1}^{N}$, the PLM $f_{\theta}(\cdot)$ and $\theta = \{\mathcal{U}, \mathcal{V}\}$, where $\mathcal{U}/\mathcal{V}$ represents the updated/frozen weights set. $\mathcal{L}(\cdot)$ indicates the loss function and $\eta$ is the learning rate.\\
    \textbf{Output}: The fine-tuned heads weight set $\mathcal{U}'$.
    
    \begin{algorithmic}[1] %[1] enables line numbers
        \STATE // \textbf{Step 1: Pre-processing}
        \STATE For each layer $l$, obtain the indicator $\delta_{h}^{l} \in \{0, 1\}$ w.r.t the $h$-th head weights set $W_{h}^{l} = \{W_{h}^{Q}, W_{h}^{K}, W_{h}^{V}\}$ by Alg. \ref{alg:info_corr} and \ref{alg:pagerank}.
        \STATE Let $W_{h}^{l} \in \mathcal{U}$ if $\delta_{h}^{l} = 1$, otherwise $W_{h}^{l} \in \mathcal{V}$.
        \STATE Freeze the parameters of $\mathcal{V}$.
        \STATE // \textbf{Step 2: Fine-tuning}
        \WHILE {not converged}
            \STATE Randomly sample the labelled data pair $(x_{i}, y_{i}) \sim \mathcal{D}$.
            \STATE Compute the loss $\mathcal{L}(\theta) = \mathcal{L}(f_{\theta}(x_{i}), y_{i})$.
            \FOR {$W_{h}^{l} \in \mathcal{U}$}
                \STATE $W_{h}^{l} := W_{h}^{l} - \eta \frac{\partial \mathcal{L}(\theta)}{\partial W_{h}^{l}}$
            \ENDFOR
        \ENDWHILE
        \STATE \textbf{return} The fine-tuned heads weight set $\mathcal{U}'$.
    \end{algorithmic}
    \label{alg:hifi}
\end{algorithm}

\begin{table*}[!h]
    \centering
    \scalebox{0.9}{
        \begin{tabular}{l | c c c c c c c c c}
            \toprule
                 & QNLI & SST-2 & $\text{MNLI}_{m}$ & $\text{MNLI}_{mm}$ & CoLA & MRPC & STS-B & RTE & QQP \\
            \midrule
                \multicolumn{10}{c}{Full-shot Learning} \\
            \midrule
                \# Train & 104,743 & 67,349 & 392,702 & 392,702 & 8,551 & 3,668 & 5,749 & 2,490 & 363,846 \\
                \# Valid & 5,463   & 872    & 9,815   & 9,796   & 1,043 & 408   & 1,500 & 277   & 40,430 \\
                \# Test  & 5,463   & 1,821  & 9,832   & 9,847   & 1,063 & 1,725 & 1,379 & 3,000 & 390,965 \\
            \midrule
                \multicolumn{10}{c}{Few-shot Learning} \\
            \midrule
                \# Train & 16$\times$2 & 16$\times$2 & 16$\times$3 & 16$\times$3 & 16$\times$2 & 16$\times$2 & - & 16$\times$2 & 16$\times$2 \\
                \# Valid & 16$\times$2 & 16$\times$2 & 16$\times$3 & 16$\times$3 & 16$\times$2 & 16$\times$2 & - & 16$\times$2 & 16$\times$2 \\
                \# Test  & 5,463     & 872       & 9,815     & 9,796     & 1,043     & 408       & - & 277       & 40,430    \\
            \bottomrule
        \end{tabular}
    }
    \caption{Statistics of each dataset on GLUE in both full-shot and few-shot scenarios.}
    \label{tab:datasets}
\end{table*}

\begin{table*}[!h]
    \centering
    \scalebox{0.8}{
        \begin{tabular}{l | c c c c c c c c c}
            \toprule
                Model & QNLI & SST-2 & $\text{MNLI}_{m}$ & $\text{MNLI}_{mm}$ & CoLA & MRPC & STS-B & RTE & QQP \\
            \midrule
                Full-FT & 32/2e-5 & 32/1e-5 & 48/2e-5 & 48/2e-5 & 32/1e-5 & 32/2e-5 & 32/2e-5 & 32/2e-5 & 32/2e-5 \\

                Diff-Pruning & 32/2e-5 & 32/5e-5 & 48/1e-5 & 48/1e-5 & 32/1e-5 & 32/1e-5 & 32/1e-5 & 32/1e-5 & 32/2e-5 \\
                
                Child-Tuning & 16/2e-5 & 16/4e-5 & 16/2e-5 & 16/2e-5 & 16/4e-5 & 16/4e-5 & 16/4e-5 & 16/4e-5 & 16/4e-5 \\
                
                Adapter & 32/3e-4 & 32/1e-4 & 48/3e-4 & 48/3e-4 & 32/3e-4 & 32/2e-4 & 32/2e-4 & 32/3e-4 & 32/3e-4 \\

                BitFit & 32/2e-4 & 32/4e-4 & 48/1e-4 & 48/1e-4 & 32/4e-4 & 32/2e-3 & 32/1e-4 & 32/1e-4 & 32/4e-4 \\

                LoRA & 32/3e-4 & 32/2e-4 & 32/2e-4 & 32/2e-4 & 32/3e-4 & 32/1e-4 & 32/3e-4 & 32/3e-4 & 32/2e-4 \\

                Compactor & 32/3e-3 & 32/3e-3 & 48/3e-3 & 48/3e-3 & 32/8e-4 & 32/2e-3 & 32/3e-3 & 32/1e-3 & 32/3e-3 \\
                
                Prefix-Tuning & 32/3e-4 & 32/3e-4 & 32/5e-4 & 32/5e-4 & 32/3e-4 & 32/2e-4 & 32/3e-4 & 32/3e-4 & 32/3e-4 \\
                
                $\text{HiFi}_{\text{layer-wise}}$ & 32/2e-4 & 32/2e-4 & 32/2e-4 & 32/2e-4 & 16/1e-4 & 16/1e-4 & 16/2e-4 & 16/2e-4 & 32/1e-4 \\

                $\text{HiFi}_{\text{mid-top}}$    & 32/1e-4 & 32/2e-4 & 32/2e-4 & 32/2e-4 & 16/3e-4 & 16/2e-4 & 16/3e-4 & 16/2e-4 & 32/2e-4 \\
            \bottomrule
        \end{tabular}
    }
    \caption{The settings of batch size / learning rate for varying baselines on a range of datasets.}
    \label{tab:settings}
\end{table*}

\subsection{More Analysis about Efficiency}
\label{appendix:efficiency}
As shown in Alg.~\ref{alg:hifi}, our approach consists of two phases: pre-processing and fine-tuning.
In the pre-processing stage, we need to calculate the information richness and correlation, where SVD is the most time-consuming operation in Alg.~\ref{alg:info_corr}. Theoretically, for a $m \times n$ matrix, the time complexity of SVD is $\mathcal{O}(n^{2} \times m + n \times m^{2})$, but we can accelerate this process by using cuSOLVER (a GPU optimization library)\footnote{\url{https://pytorch.org/docs/stable/generated/torch.svd.html}} on CUDA. Besides, the correlation can be solved almost in linear time. For the PageRank algorithm in Alg.~\ref{alg:pagerank}, the theoretical time complexity is $\mathcal{O}(t(\epsilon) \times n^{2})$, where $t(\epsilon)$ and $n$ refer to the number of iterations and nodes, respectively. As shown in the right of Fig.~\ref{fig:effect_head_damping}, $t(\epsilon) < 12$ under diverse datasets, which indicates the fast convergence of PageRank.

Overall, our pre-processing process is so efficient that the time spent on it is almost negligible compared to fine-tuning. Once the heads to be updated are determined, the fine-tuning efficiency of our method HiFi is similar to other methods, e.g., BitFit, Child-Tuning.

\subsection{More Experimental Results on Commonly-used Tasks}
In addition to evaluating the natural language understanding tasks in the main text, we also conduct the comparison on two widely-used benchmarks: SQuAD \cite{rajpurkar2016squad} and SWAG \cite{zellers2018swag}. For each method, we train only $3$ epochs on the training set and then report the result on the validation set are shown in Tab.~\ref{tab:more_res}.

\begin{table}[h]
    \centering
    \scalebox{0.9}{
        \begin{tabular}{l | c c}
            \toprule
                Method & SQuAD (F1) & SWAG (Acc.) \\
            \midrule
                Full-FT       & 90.7 & 85.5 \\
            \midrule
                Diff-Pruning  & 89.0 & 84.2 \\
                Child-Tuning  & 88.5 & 83.7 \\
            \midrule
                Adapter       & 88.0 & 83.8 \\
                BitFit        & 82.1 & 81.4 \\
                LoRA          & 89.2 & 83.1 \\
                Compacter     & 87.1 & 83.8 \\
                Prefix-tuning & 87.7 & 84.3 \\
                HiFi          & 89.8 & 84.9 \\
            \bottomrule
        \end{tabular}
    }
    \caption{The experimental results on the validation datasets of SQuAD and SWAG.}
    \label{tab:more_res}
\end{table}

\end{document}